\documentclass{bmvc2k}
\usepackage{amsmath}
\usepackage{amssymb}
\usepackage{breqn}
\usepackage{mwe}
\usepackage[font=scriptsize,labelfont=bf]{caption}
\usepackage{color,soul}
\usepackage[numbers]{bmvc2k_natbib}
\usepackage{graphicx}
\usepackage{subfigure}
\usepackage[export]{adjustbox}
\usepackage[normalem]{ulem}
\usepackage{wrapfig,lipsum,booktabs}
\usepackage{blindtext}

\setstcolor{red}

\title{Adaptive Noise-Tolerant Network for Image Segmentation}

\runninghead{Student, Prof, Collaborator}{BMVC Author Guidelines}

\newcommand{\etal}{et~al\mbox{.}}

\usepackage{titling} 
\usepackage{authblk}
\author{Weizhi Li\footnote{Work was done by the author while affiliated  with Texas A\&M University in 2017. }}

\begin{document}
\date{} 
\maketitle

\begin{abstract}
Unlike image classification and annotation, for which deep network models have achieved dominating superior performances compared to traditional computer vision algorithms, deep learning for automatic image segmentation still faces critical challenges. One of such hurdles is to obtain ground-truth segmentations as the training labels for deep network training. Especially when we study biomedical images, such as histopathological images (histo-images), it is unrealistic to ask for manual segmentation labels as the ground truth for training due to the fine image resolution as well as the large image size and complexity. In this paper, instead of relying on clean segmentation labels, we study whether and how integrating imperfect or noisy segmentation results from off-the-shelf segmentation algorithms may help achieve better segmentation results through a new Adaptive
Noise-Tolerant Network (ANTN) model. We extend the noisy label deep learning to image segmentation with two novel aspects: (1) multiple noisy labels can be integrated into one deep learning model; (2) noisy segmentation modeling, including probabilistic parameters, is adaptive, depending on the given testing image appearance. Implementation of the new ANTN model on both the synthetic data and real-world histo-images demonstrates its effectiveness and superiority over off-the-shelf and other existing deep-learning-based image segmentation algorithms.

\end{abstract}

\section{Introduction}
\label{sec:intro}

Many deep learning models operate in a supervised nature and have had enormous successes in many applications, including image classification and annotation~\cite{mnih2012learning, xiao2015learning,veit2017learning,sukhbaatar2014training,reed2014training,kakar2015if}. However, when we have only unlabeled data, for example, for biomedical image analysis, the existing Convolutional Neural Network (CNN) based methods may not apply. In order to overcome such limitations, a crowd-sourcing strategy~\cite{albarqouni2016aggnet} has been proposed to collect manual annotations from different people when getting labels from medical experts is difficult and then integrate the cues from these noisy crowd-sourcing annotations to infer the ground truth~\cite{lakshminarayanan2013inferring}. In image classification, these noisy crowd-sourcing labels are considered to be Noisy At Random (NAR)~\cite{frenay2014classification} and several NAR models~\cite{sukhbaatar2014training, mnih2012learning} have been proposed by modeling the label-flip noise independent of input images. 

In this paper, we focus on deep learning for image segmentation when it is difficult to obtain clean pixel-wise segmentation labels. There are indeed many existing traditional image segmentation algorithms that can provide segmentation results with reasonable quality and can be taken as noisy training labels. However, for these segmentation labels, the NAR assumption may not hold any more, since the label-flip noise is not only dependent on the true object class but also the image features of the corresponding classes~\cite{frenay2014classification}. Hence, it is more proper to consider the Noisy Not At Random (NNAR) model~\cite{frenay2014classification} and develop adaptive image-dependent label-flip noise transition models. Motivated by the recent dynamic filter networks~\cite{de2016dynamic} that adaptively adjust deep network parameters accordingly to the input features, we propose an Adaptive Noise-Tolerant Network (ANTN) for image segmentation. The graphical probabilistic model and architecture of the network are shown in Figure 1. In ANTN, we explicitly model the probabilistic dependency between the input image, the ground-truth segmentation, and noisy segmentation results from off-the-shelf image segmentation algorithms. By adaptively modeling image-dependent label-flip noise from different segmentation algorithms, ANTN can borrow signal strengths from multiple noisy labels to achieve better segmentation results. 

We develop an EM (Expectation-Maximization) based model inference algorithm and apply ANTN for image segmentation with both synthetic and histo-images. Performance comparison with off-the-shelf and deep learning algorithms shows the effectiveness and superiority of ANTN over other competing algorithms.

\begin{figure}[h]
\begin{minipage}{1.1cm}
\centering
\includegraphics[width=1.1cm]{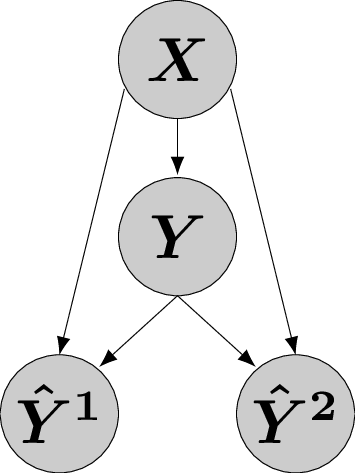}\\\centering\scriptsize{(a)}
\end{minipage}
\hfill
\centering
\begin{minipage}{11.7cm}
\centering
\includegraphics[width=11.7cm]{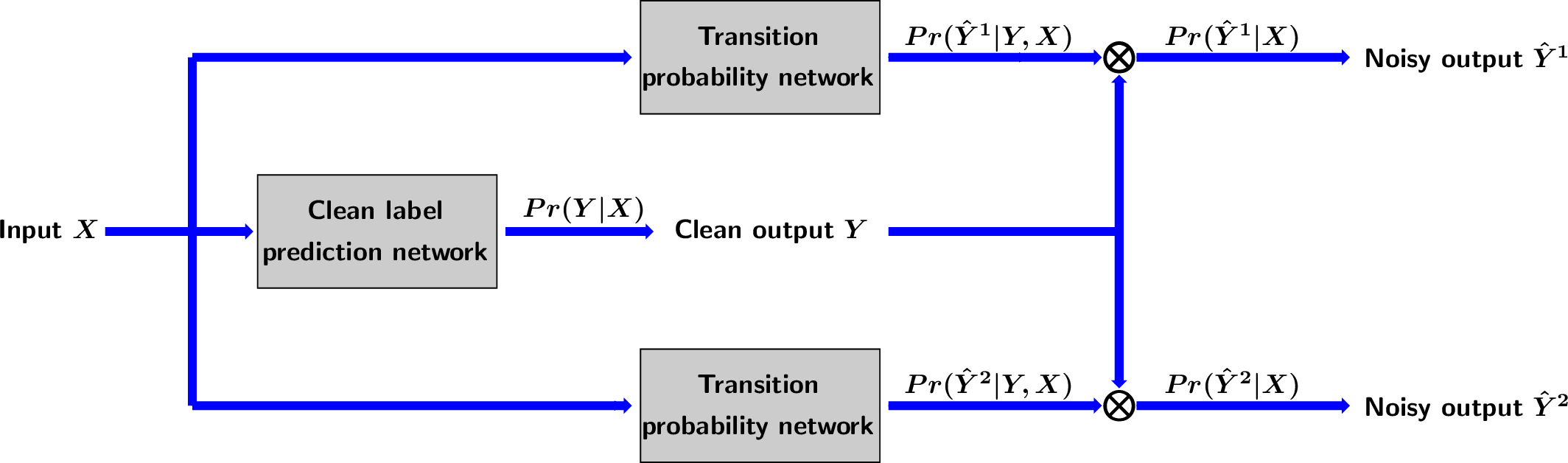}\\\centering\scriptsize{(b)}
\end{minipage}\vspace{0.2cm}
\caption{(a) Graphical probabilistic model and (b) architecture of the Adaptive Noise-Tolerant Network (ANTN). $X$ represents the input image, $Y$ represents the ground-truth segmentation, $\hat{Y}^1$ and $\hat{Y}^2$ represent noisy segmentations.}
\vspace{-0.1in}
\end{figure}

\vspace{0.0001mm}
\section{Related Work}
\label{sec:relawork}
Hinton~\etal~\cite{mnih2012learning} have pioneered to use the deep network to incorporate the label-flip noise in aerial image labeling. They assumed the label-flip noise is only dependent on the true label and adopted an EM algorithm to train network model parameters iteratively, considering the true labels as latent variables. Beno{\^\i}t~\etal~\cite{frenay2014classification} pointed out it is more realistic that mislabeling is dependent on input features and Xiao~\etal~\cite{xiao2015learning} took such an assumption into consideration and integrated three types of label noise transition probabilities given the same true label for clothing classification. Instead of modeling true labels as latent variables in~\cite{mnih2012learning, xiao2015learning}, Veit~\etal~\cite{veit2017learning} recently introduced a multi-task label cleaning architecture for image classification, in which an image classifier is supervised by the output of the label cleaning network trained using the mixture of clean and noisy labels, which is effective in learning large-scale noisy data in conjunction with a small subset of clean data.  

All the aforementioned methods~\cite{mnih2012learning,xiao2015learning, veit2017learning} require a small clean dataset to assist model inference. Sukhbaatar~\etal~\cite{sukhbaatar2014training} proposed a noisy label image classification model that is capable of learning network parameters from noisy labels solely by diffusing label noise transition probability matrix from the initial identity matrix with a weight decay trick; but such model inference can be instable. Reed~\etal~\cite{reed2014training} introduced ``prediction consistency'' in training a feed-forward autoencoder with noisy data, requiring that the same label prediction should be made given similar image features. Similar to the idea of avoid overfitting, Kakar~\etal~\cite{kakar2015if} added a regularization term on the coefficients of hidden units during training to obtain stable results. 


Most of the existing noisy label deep network models~\cite{mnih2012learning, xiao2015learning,veit2017learning,sukhbaatar2014training,reed2014training,kakar2015if} are on image classification or patch-labeling. The authors in~\cite{li2017noise} extended the label-flip noise model~\cite{sukhbaatar2014training} to image segmentation and devised a Noise-Tolerant Network (NTN) based on the architecture of u-net~\cite{ronneberger2015u} assuming that the label noise is only dependent on the true label. For image segmentation, it may be more appropriate to model label transition probabilities adaptively as discussed earlier. In addition, when different types of noisy labels are available, there is still no existing method to flexibly integrate them in the literature to the best of our knowledge. Our ANTN specifically addresses these two problems for deep-learning image segmentation. 

\vspace{-0.1in}
\section{Method}
\label{sec:Method}

We study whether and how integrating multiple noisy datasets from off-the-shelf segmentation algorithms may help achieve better segmentation results.
We develop the ANTN model for image segmentation with two novel aspects: (1) multiple noisy labels can be naturally integrated; (2) noisy segmentation modeling is adaptive, with image-dependent label transition probabilities. 
First, we introduce the probabilistic model of image segmentation when multiple noisy segmentations are available. Based on the model, we construct the adaptive deep learning framework, motivated by the recent dynamic filter network~\cite{de2016dynamic}. We note that we focus on the model with two noisy datasets but the model can be generalized to the settings with more than two noisy datasets in a straightforward manner due to the symmetry of our proposed framework. We also provide the model inference procedure to train the proposed adaptive deep learning model. 

\vspace{-0.1in}
\subsection{Model}

Given a set of training images $X=\{X_1, X_2, \ldots, X_T\}$, which could be sub-images or  patches, we can apply $S$ selected off-the-shelf segmentation algorithms to obtain noisy or imperfect segmentations $\hat{Y}^1 = \{\hat{Y}_1^1, \hat{Y}_2^1, \ldots, \hat{Y}_T^1\}$, $\hat{Y}^2$, $\ldots$, $\hat{Y}^S$. To clearly convey the idea, we focus on the settings with $S=2$ in this paper. We can model the relationships between input images and noisy segmentations based on the following general probabilistic model:

\begin{equation}
Pr\big(\hat{Y}^1, \hat{Y}^2|X\big) = \sum_{Y \in C^{|\mathcal{I}|}}Pr\big(\hat{Y}^1, \hat{Y}^2, Y|X\big)=\sum_{Y \in C^{|\mathcal{I}|}}Pr\big(\hat{Y}^1|Y,X\big)Pr\big(\hat{Y}^2|Y,X\big)Pr\big(Y|X\big), \label{eq:jml}
\end{equation}
in which $C$ is the total number of label classes for segmentation; $Y$ denotes the clean or perfect segmentations; and $|\mathcal{I}|$ represents the total number of pixels in $X$ indexed by the pixel set $\mathcal{I}$. We note that in~\cite{mnih2012learning,sukhbaatar2014training,li2017noise}, (1) the clean pixel-wise labels indexed by $n$: ${y}_n$'s, are conditionally independent given $X$: $Pr\big(Y|X\big) = \Pi_{n\in \mathcal{I}} Pr(y_n|X)$; and (2) the noisy pixel-wise labels $\hat{y}_n$'s are conditionally independent with $X$ given $Y$ and the pixel-wise label transition probabilities are identical: $Pr\big(\hat{Y}|Y,X\big) =\Pi_{n\in \mathcal{I}} Pr(\hat{y}_n|y_n)$. Hence, the log-likelihood with one set of noisy labels for $X$ can be written as:
\begin{align}
L = \log{Pr\big(\hat{Y}|X\big)} &= \sum_{n\in \mathcal{I}} \log{\Big[\sum_{y_n=1}^{C}Pr\big(\hat{y}_n|y_n\big)Pr\big(y_n|X\big)\Big]},
\end{align}
with which a Noise-Tolerant Network (NTN) with the u-net architecture~\cite{li2017noise} has been developed to recover clean segmentations. 

In the proposed ANTN model (1), we relax the second assumption in NTN when integrating multiple types of noisy labels. For different pixels, the transition probability of the noisy label given the clean label will be dependent on $X$ since segmentation results from different algorithms can be dependent on both images and segmentation algorithms. Let $Pr_n(\hat{y}_n^s|y_n,X)$ denote the new transition probabilities, where $s=1$, $2$ for different noisy segmentations. Following the dynamic filter network models in~\cite{de2016dynamic}, we can rewrite the probabilistic model (1): 
\begin{equation}
Pr\big(\hat{Y}^1, \hat{Y}^2|X\big) = \prod_{n\in \mathcal{I}} \sum_{y_n=1}^{C}Pr_n(\hat{y}_n^1|y_n,X)Pr_n(\hat{y}_n^2|y_n,X)Pr(y_n|X). \label{eq:rjml}
\end{equation}
With this model, we can construct respective deep learning models for all the involved probability distribution functions, including the clean label probability $Pr(Y|X)$, pixel-wise conditional probabilities $Pr_n(\hat{y}_n^1|y_n,X)$ and $Pr_n(\hat{y}_n^2|y_n,X)$, as illustrated by the schematic graphical model for recovering clean labels from two noisy datasets in Figure 1(a). We note the symmetry of the proposed deep learning framework, which enables the straightforward generalization when $S>2$. For each of the three components in Figure 1(a), we follow the construction in~\cite{ronneberger2015u,mnih2012learning,li2017noise,de2016dynamic} to have the corresponding u-net architectures with the deep network framework shown in Figure 1(b). 
The main difference among these three deep network models are the constraints applied to their outputs of the last layers: 
\begin{align}
\sum_{y_n=1}^{C}Pr_n(y_n|X)=1,\quad\sum_{\hat{y}_n^1=1}^{C}Pr_n(\hat{y}_n^1|y_n,X)=1,\quad\sum_{\hat{y}_n^2}^{C}Pr_n(\hat{y}_n^2|y_n,X)=1,
\end{align}
which guarantee the legitimacy of the modeled probability distribution functions.

We note that the clean label model $Pr(Y|X)$ has to be combined with the noise transition network models $Pr_n(\hat{y}_n^1|y_n,X)$ and $Pr_n(\hat{y}_n^2|y_n,X)$ for training as we do not observe the ground-truth segmentations. The integration of the three components in~Figure 1(a) is motivated by the label-flip noise model in the noise-tolerant image classification framework in~\cite{sukhbaatar2014training} and the introduced asymmetric Bernoulli noise~(ABN) model in~\cite{mnih2012learning}. 

\vspace{-0.1in}
\subsection{Model Inference}

Due to the unobserved clean segmentation labels, training three different components given $X$ and noisy segmentations $\hat{Y}^1$ and $\hat{Y}^2$ is an iterative procedure to maximize the following three log-likelihood functions based on the model (3): 
\begin{align}
\mathcal{L}_s &= \frac{1}{N}\sum_{n\in\mathcal{I}}log\sum_{y_n=1}^{C}Pr((\hat{y}_n^s)_{obs}|y_n,X;\theta_s)Pr(y_n|X;\theta_3),\quad s=1,2, \\
\mathcal{L}_3 &= \frac{1}{N}\sum_{n\in\mathcal{I}}log\sum_{y_n=1}^{C}Pr((\hat{y}_n^1)_{obs}|y_n,X;\theta_1)Pr((\hat{y}_n^2)_{obs}|y_n,X;\theta_2)Pr(y_n|X;\theta_3)
\end{align}
where $\theta_1$, $\theta_2$ and $\theta_3$ are the corresponding network parameters of two transition probability networks and the clean label prediction network; $(\hat{y}_n^1)_{obs}$ and $(\hat{y}_n^2)_{obs}$ denote observed noisy labels; and $N=|\mathcal{I}|$. We alternate the order of optimization with respect to $\theta_1$ and $\theta_2$ for minimizing (5) and $\theta_3$ for minimizing (6). Similar to~\cite{mnih2012learning,sukhbaatar2014training}, we consider $Y$ as latent variables and maximize the likelihood functions by the EM algorithm:

{\raggedleft\textbf{E-step:}}
Given deep network paramters $\theta_1^{(t)}$, $\theta_2^{(t)}$ and $\theta_3^{(t)}$ for three component networks at each iteration, the posterior probabilities of the latent segmentation label $Pr(y_n|(\hat{y}_n^s)_{obs}, X)$ and $Pr(y_n|(\hat{y}_n^1)_{obs},(\hat{y}_n^2)_{obs}, X)$ for the corresponding likelihood functions (5) and (6) can be updated as follows:
\begin{align}
Pr^{(t)}(y_n|(\hat{y}_n^s)_{obs}, X)&= \frac{Pr((\hat{y}_n^s)_{obs}|y_n, X;\theta_s^{(t)})Pr(y_n|X; \theta_3^{(t)})}{\sum_{y_n=1}^CPr(\hat{y}^s_{obs}|y_n, X;\theta_s^{(t)})Pr(y_n|X; \theta_3^{(t)})},\quad s=1,2,\notag\\
Pr^{(t)}(y_n|(\hat{y}_n^1)_{obs}, (\hat{y}_n^2)_{obs}, X)&= \frac{Pr((\hat{y}_n^1)_{obs}|y_n, X;\theta_1^{(t)})Pr((\hat{y}_n^2)_{obs}|y_n, X;\theta_2^{(t)})Pr(y_n|X;\theta_3^{(t)})}{\sum_{y_n=1}^CPr((\hat{y}_n^1)_{obs}|y_n, X; \theta_1^{(t)})Pr((\hat{y}_n^2)_{obs}|y_n, X; \theta_2^{(t)})Pr(y_n|X;\theta_3^{(t)})}.\notag
\end{align}
{\raggedleft\textbf{M-step:}} With the estimated posterior probabilities, we update the corresponding network parameters through optimizing the expected complete likelihood functions. In practice, we cannot guarantee the optimality of M-step updates due to our deep network modeling. We implement gradient descent and backprogation in the corresponding component networks to update parameters as follows: 
\begin{align}
\triangledown\theta_s^{(t+1)} &\leftarrow \frac{1}{N}\sum_{n\in\mathcal{I}}\sum_{y_n=1}^CPr^{(t)}(y_n|(\hat{y}_n^s)_{obs},X)\frac{\partial{logPr((\hat{y}_n^s)_{obs}|y_n,X;\theta_s)}}{\partial \theta_s},\quad s=1,2,\\
\triangledown\theta_3^{(t+1)} &\leftarrow \frac{1}{N}\sum_{n\in\mathcal{I}}\sum_{y_n=1}^CPr^{(t)}(y_n|(\hat{y}_n^1)_{obs}, (\hat{y}_n^2)_{obs},X)\frac{\partial{logPr(y_n|X;\theta_3)}}{\partial \theta_3}.
 \end{align}
For transition probability networks, we only observe one noisy label for each pixel and we can only  unambiguously derive $Pr((\hat{y}_n^s)_{obs}|y_n,X; \theta_s)$. For the other transition probabilities, we simply set them to be $[1 - Pr((\hat{y}_n^s)_{obs}|y_n,X; \theta_s)]/(C - 1)$.

For the complete procedure of ANTN model inference, we first initialize the clean label prediction network by training with the mixture of noisy datasets, then train each transition probability network with the corresponding noisy labels as described in the EM algorithm. After these two steps, we iteratively train the component networks by alternating the optimization with a fixed number of interval epochs for each of them  until convergence. 

\vspace{-0.1in}
\section{Experimental Results}

We validate the effectiveness of ANTN by comparing it with off-the-shelf and deep-learning image segmentation algorithms on both synthetic and histo-images. 

\vspace{-0.1in}
\subsection{Datasets}

To quantitatively evaluate ANTN and compare it with other segmentation algorithms, we first create a synthetic image set with the corresponding simulated noisy segmentations. With the validated performance improvement over existing algorithms, we then apply ANTN to a set of histo-images, obtained from a study of Duchenne Muscular Dystrophy (DMD) disease~\cite{klingler2012role}, for performance evaluation. 

\textbf{Synthetic Data:} 
We generate 135 $472\times472$ synthetic images for quantitative performance evaluation. First, we randomly simulate red, green, and blue circular objects with different radii uniformly distributed from 15 to 40 pixels in each image. Hence, there are four classes required to be segmented: red, green, and blur circular objects as well as white background regions. For each of RGB channels, the corresponding intensities for pixels in each class follow a Gaussian distribution with the mean 200 and standard variation 50. An example of the generated synthetic images and the corresponding ground truth for its object segmentation are shown in Figures 2(a) and~(c). To further create different types of noisy segmentation labels, we erode and dilate the ground-truth segmentation by a rectangle structural element with the width and length set to 5 pixels, with the generated noisy labels given in Figures 2(b) and (d) for the corresponding image example.


\textbf{Histopathological Images:} 
We also have obtained 11 samples of ultra-high resolution histo-images for studying DMD~\cite{klingler2012role}. They are split into $472\times472$ sub-images and preprocessed by a stain normalization method~\cite{reinhard2001color}. Some of the preprocessed sub-images are shown in Figure 5(a). For these images, we are interested in quantifying the percentage of fibrosis (stained blue) and muscle (stained pink) to estimate the seriousness of the disease~\cite{klingler2012role,gurcan2009histopathological}. Hence, the segmentation task is to segment fibrosis (blue), muscle (pink), and other tissue types (white). We have applied two simple off-the-shelf segmentation algorithms: K-Means~\cite{hartigan1979algorithm} and Otsu thresholding~\cite{otsu1975threshold} on all the sub-images and we consider the obtained segmentation labels as the noisy segmentation labels in deep-learning methods including ANTN. Segmentation examples are shown in Figures 5(b) and (c).  


\vspace{-0.1in}
\subsection{Performance evaluation on synthetic data}

\begin{wrapfigure}{r}{0.5\textwidth}
\vspace{-0.15in}
\centering
\begin{minipage}{1.45cm}
\centering
\includegraphics[width=1.45cm]{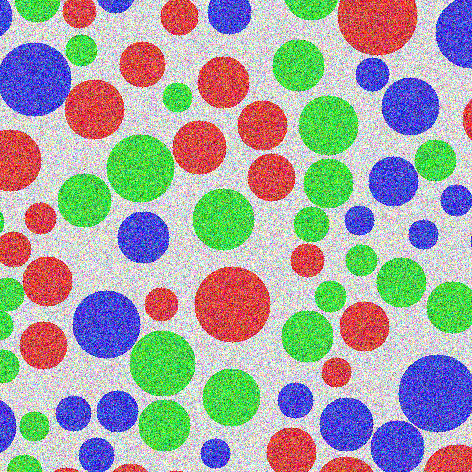}\\\centering\scriptsize{(a) Original}\vspace{0.05cm}
\end{minipage}\hspace{0.07cm}
\centering
\begin{minipage}{1.45cm}
\centering
\includegraphics[width=1.45cm]{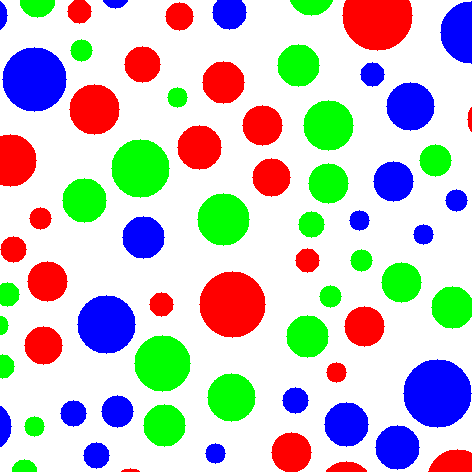}\\\centering\scriptsize{(b) Erosion}\vspace{0.07cm}
\includegraphics[width=1.45cm]{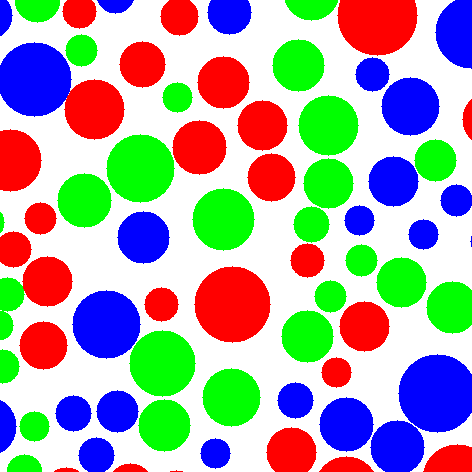}\\\centering\scriptsize{(c) Clean label}\vspace{0.07cm}
\includegraphics[width=1.45cm]{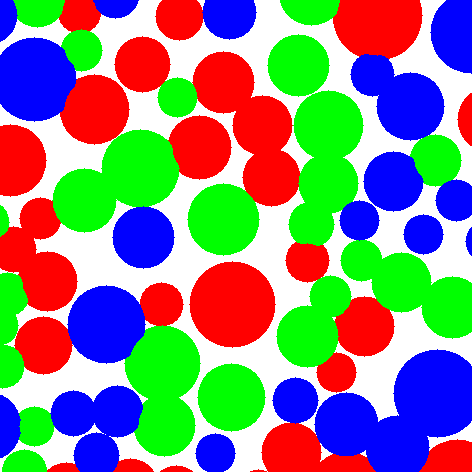}\\\centering\scriptsize{(d) Dilation}
\end{minipage}\hspace{0.07cm}
\centering
\begin{minipage}{1.45cm}
\centering
\includegraphics[width=1.45cm]{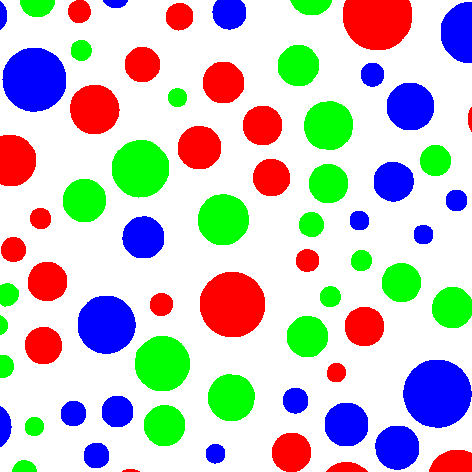}\\\centering\scriptsize{(e) U-net $^1$}\vspace{0.07cm}
\includegraphics[width=1.45cm]{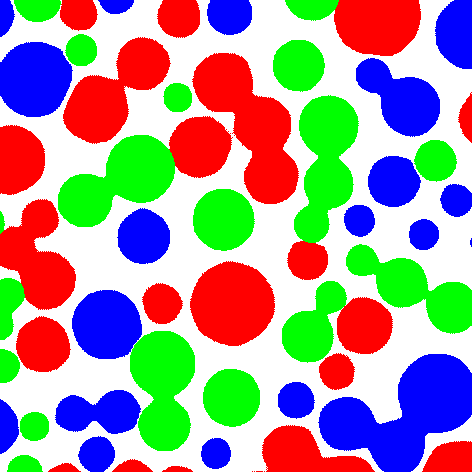}\\\centering\scriptsize{(f) U-net$^3$}\vspace{0.07cm}
\includegraphics[width=1.45cm]{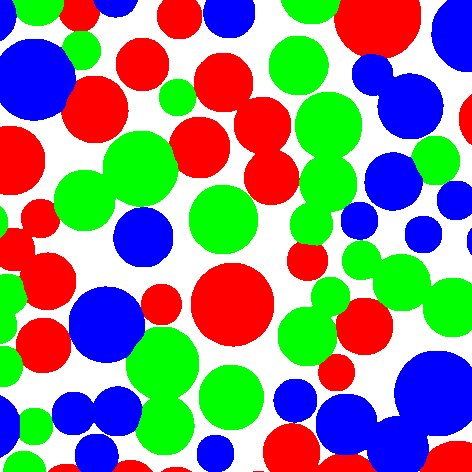}\\\centering\scriptsize{(g) U-net$^2$}
\end{minipage}\hspace{0.07cm}
\centering
\begin{minipage}{1.45cm}
\centering
\includegraphics[width=1.45cm]{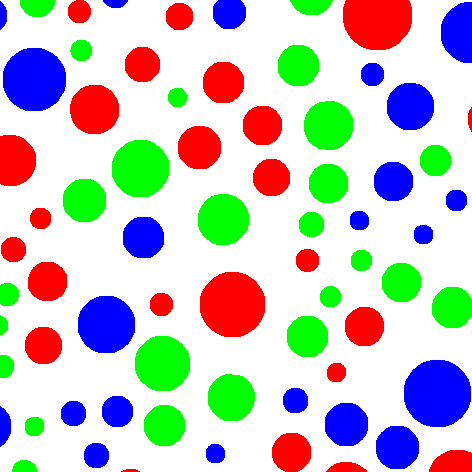}\\\centering\scriptsize{(h) NTN$^1$}\vspace{0.07cm}
\includegraphics[width=1.45cm]{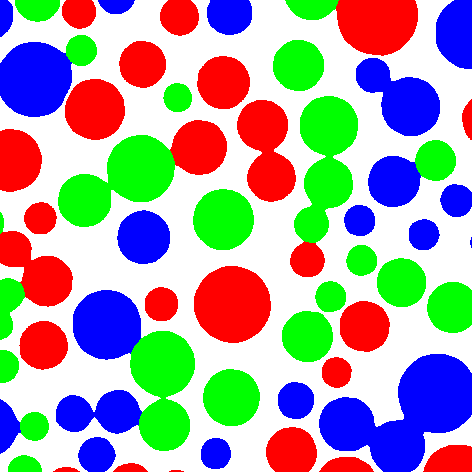}\\\centering\scriptsize{(i) ANTN}\vspace{0.07cm}
\includegraphics[width=1.45cm]{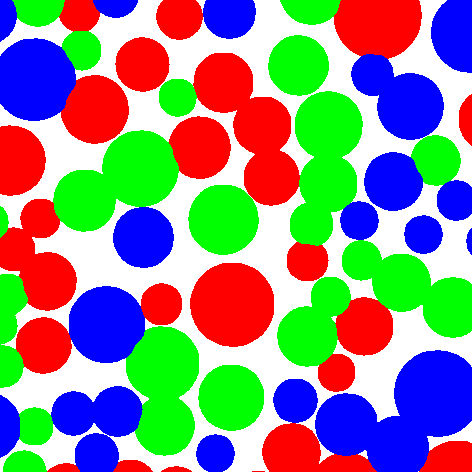}\\\centering\scriptsize{(j) NTN$^2$}
\end{minipage}\vspace{0.3cm}
\caption{Synthetic image and corresponding segmentations.}
\end{wrapfigure}

For synthetic data, we compare the performance of ANTN with the existing deep-learning segmentation methods: (1) U-net~\cite{ronneberger2015u} that is a CNN taking noisy segmentation labels as the ground truth for training; (2) Noise-tolerant u-net~(NTN)~\cite{sukhbaatar2014training,li2017noise} that models the segmentation noise independent of image features. 35 synthetic images and their corresponding segmentations are used for training.
For our ANTN, we first initialize the clean label prediction network (a u-net with the same architecture as in~\cite{ronneberger2015u,li2017noise}) by training with a mixture of two noisy datasets for the first 100 epochs, then train both transition probability networks (two similar u-nets) by the proposed EM-algorithm with the corresponding erosion and dilation noisy segmentations in next 200 epochs. Finally, we iteratively train the whole network setting the alternating interval to be 10 epochs for next 200 epochs. We keep the learning rate at $10^{-4}$ for the first 450 epochs and $10^{-5}$ for the last 50 epochs. For competing methods, we directly train the u-net considering either erosion, dilation, or their mixture as the ground-truth segmentation. With erosion and dilation noisy labels, the training procedure converges for 200 epochs. With the mixture of noisy labels, it converges for 100 epochs. For NTN, in addition to training the original u-net layers, we also train the label-flip-noise transition layer with the corresponding noisy labels by weight decay to diffuse the label-flip-noise transition probability from identity to approximate the average noise transition probability matrix for 150 more epochs~\cite{sukhbaatar2014training, li2017noise}.
We do not train NTN with the mixture of noisy labels as it can only take one single type of noisy labels~\cite{sukhbaatar2014training,li2017noise}. Training of the u-net with different noisy labels can be considered as the intermediate steps of ANTN and NTN model inference. 



We provide the examples of the corresponding segmentation results in Figures 2(e)-(j), in which u-net$^1$, u-net$^2$, and u-net$^3$ represent the u-nets trained with the corresponding erosion, dilation, and mixture of noisy segmentations; NTN$^1$ and NTN$^2$ represent the NTNs trained with the corresponding erosion and dilation noisy segmentions. It is clearly that the u-net or NTN~\cite{sukhbaatar2014training,li2017noise} often can not correctly segment the corresponding objects without appropriate modeling of segmentation noise with erosion and dilation bias.
In Figures 2(i), it is clear that our ANTN performs the best due to the adaptive integration of label-flip-noise transitions. In addition, the performance improvement may also come from the integration of multiple types of noisy labels with the capability of borrowing signal strengths. We further quantitatively evaluate segmentation accuracy by the synthetic test dataset of 100 images and get the highest accuracy of \textbf{97.71\%} for ANTN followed by \textbf{93.38\%} by u-net$^3$ with mixture training being superior over other methods, which have obained the accuracy all below \textbf{85\%}.







In order to show the convergence of our training procedure, we analyze the trends of the cross entropy between the intermediate segmentation labels during training and the clean ground-truth labels, as well as the noisy labels taken for training. 
\begin{figure}
\centering
\begin{minipage}{3.1cm}
\centering
\includegraphics[width=3.1cm]{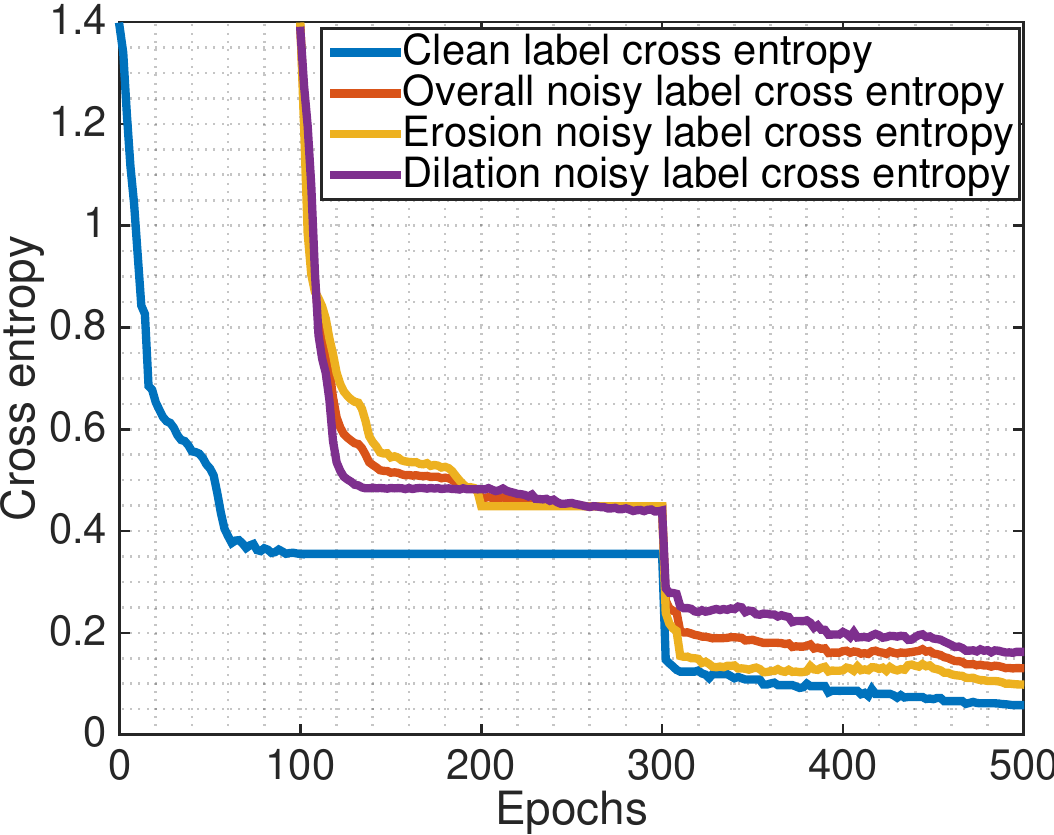}\\\centering\scriptsize{(a)}
\end{minipage}
\centering
\begin{minipage}{3.1cm}
\centering
\includegraphics[width=3.1cm]{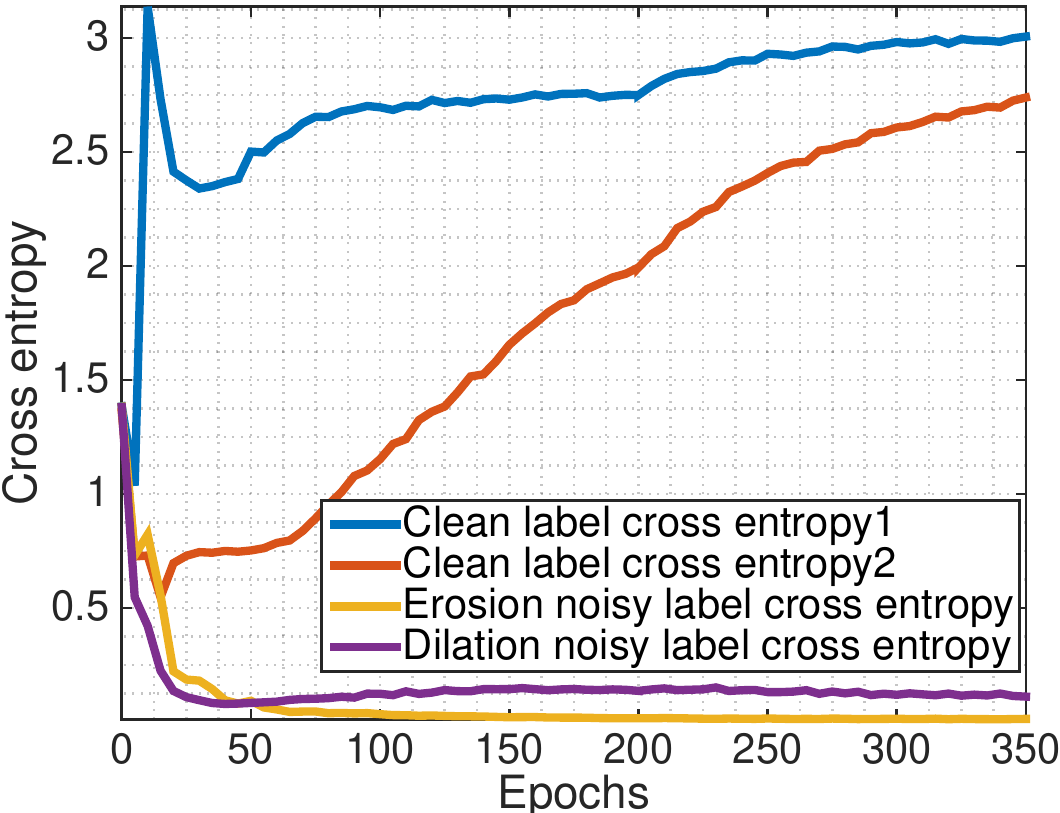}\\\hspace{0.28cm}\centering\scriptsize{(b)}
\end{minipage}
\centering
\begin{minipage}{3.1cm}
\centering
\includegraphics[width=3.1cm,height=2.5cm]{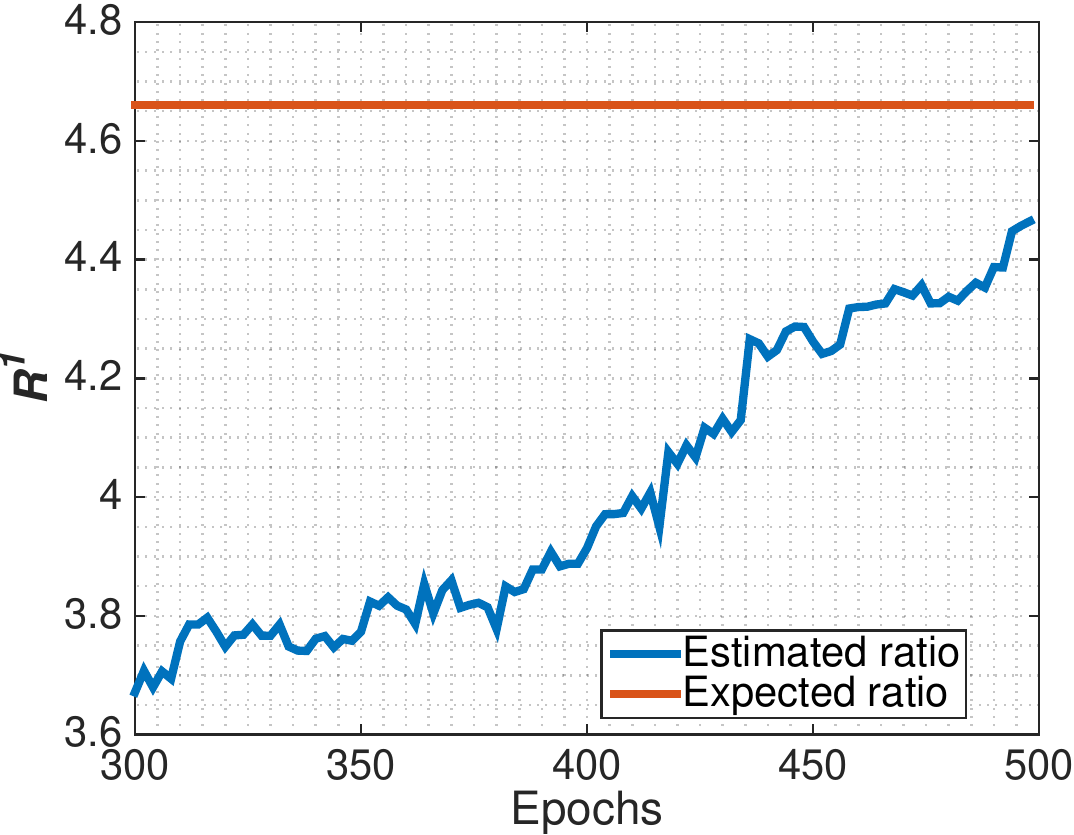}\\\hspace{0.3cm}\centering\scriptsize{(c)}
\end{minipage}
\centering
\begin{minipage}{3.1cm}
\centering
\includegraphics[width=3.1cm,height=2.5cm]{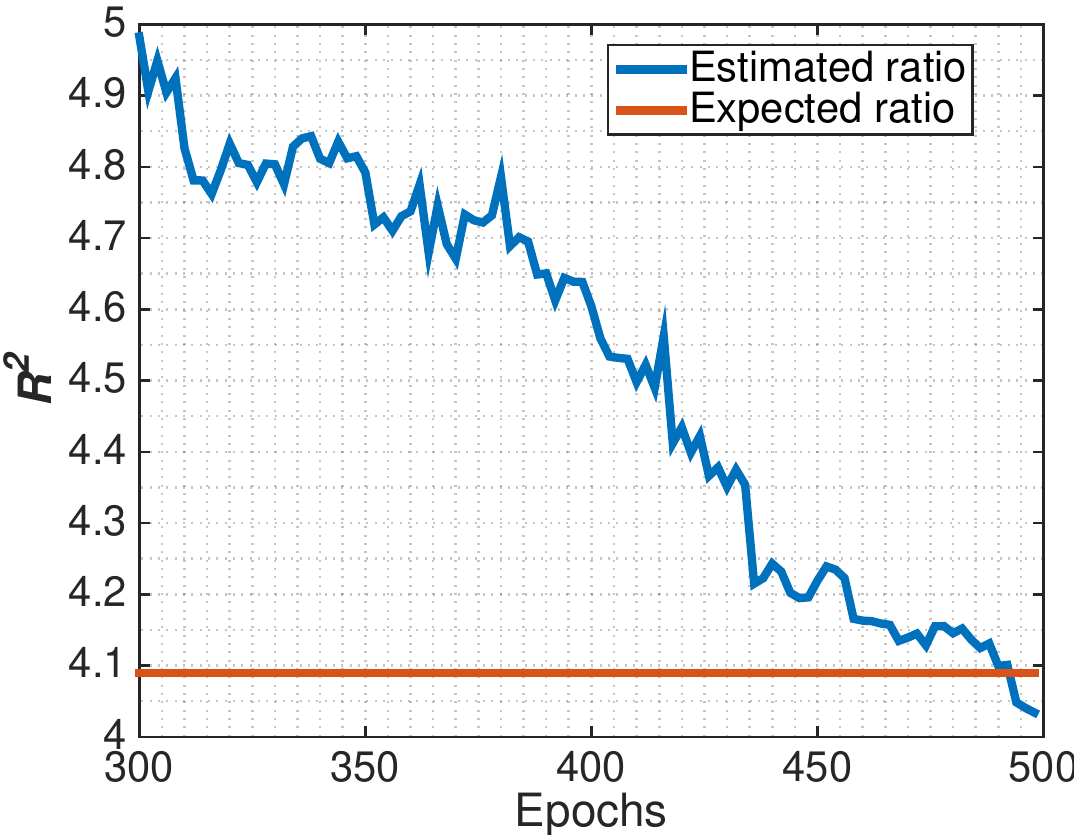}\\\hspace{0.3cm}\centering\scriptsize{(d)}
\end{minipage}
\vspace{0.2cm}
\caption{(a)~Cross entropy evaluation for the u-net$^3$ and ANTN.~(b)~Cross entropy evaluation for the u-net$^1$,u-net$^2$ and NTNs. (c)~Estimated clean-label ratio for erosion dataset.~(d)~Estimated clean-label ratio for dilation dataset}
\end{figure}
From Figure 3(a), we observe that the training of the clean label network in ANTN converges around 100 epochs with the clean-label
cross entropy reaching the plateau. Note that the intermediate results at this point is also the final results of u-net$^3$ training with the mixture of noisy labels. After that, we implement EM algorithm to train two noise transition probability networks. Clearly, the change of the noisy-label cross entropy indicates that the training of two transition probability networks converges in the next 200 epochs. During the next iterative training procedure, we observe the corresponding cross entropy values drop drastically and then continuously decrease till convergence. Figure 3(b) shows the corresponding cross entropy changes during u-net as well as NTN training with either erosion or dilation noisy datasets. The training for u-net stops at 200 epochs which also serves the initialization of NTN training before the noise transition layer training. We can see that the clean-label cross entropy diverges gradually though the noisy-label cross entropy decreases till convergence. This is because no component in u-net models potential segmentation noise. 


To further validate the convergence and effectiveness of ANTN, we compare the ratio $R$ of the estimated clean labels to the corresponding $s$th type of noisy labels during training with the actual ratio of clean labels to noisy labels for the corresponding erosion or dilation training outputs, as shown in Figures 3(c) and (d): 
\begin{align}
R^s=\frac{\sum_{n\in\mathcal{I}}I(\operatorname*{arg\,max}_uPr(y_n=u|(\hat{y}^1_n)_{obs},(\hat{y}^2_n)_{obs},X)=(\hat{y}^s_n)_{obs})}{\sum_{n\in\mathcal{I}}I(\operatorname*{arg\,max}_uPr(y_n=u|(\hat{y}^1_n)_{obs},(\hat{y}^2_n)_{obs},X)\neq(\hat{y}^s_n)_{obs})},\quad s=1,2 .
\end{align}
From Figures 3(c) and (d), the estimated ratios indeed approach the actual ratios in the training data with the corresponding trend indicating the learned ANTN models the noise transitions better and better during the iterative training stage. 


Finally, we check the noisy transition matrices learned by NTN and the average transition matrices for ANTN, compared to the expected noisy transition matrices obtained by clean and noisy training data. We emphasize that the noisy transition matrix in ANTN is pixel-wise and dependent on image features, we compute the average transition matrices by simply averaging pixel-wise transition probabilities across training images. Clearly, ANTN can better approximate the noise transition by visual comparison in Figure 4. 

\setlength{\belowcaptionskip}{-13pt}
\begin{wrapfigure}{r}{0.38\textwidth}
\vspace{-0.1in}
\centering
\begin{tabular}{@{}c@{}c@{}c@{}}
\includegraphics[width=1.65cm]{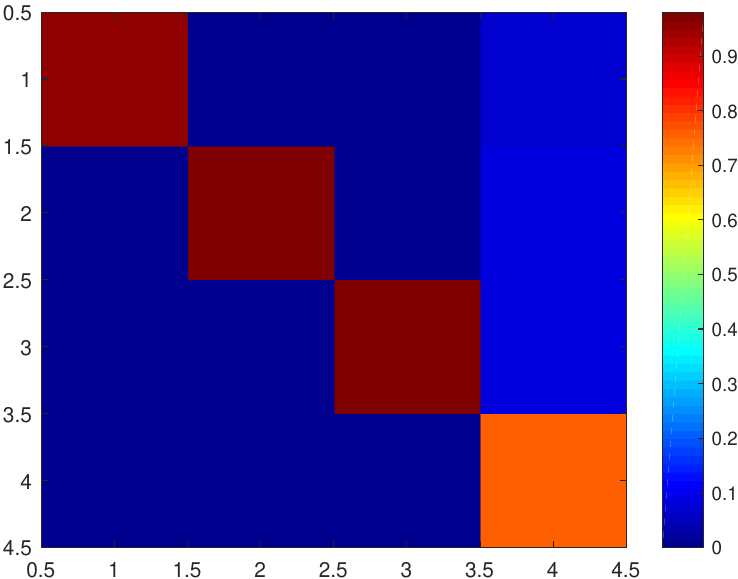}&
\includegraphics[width=1.65cm]{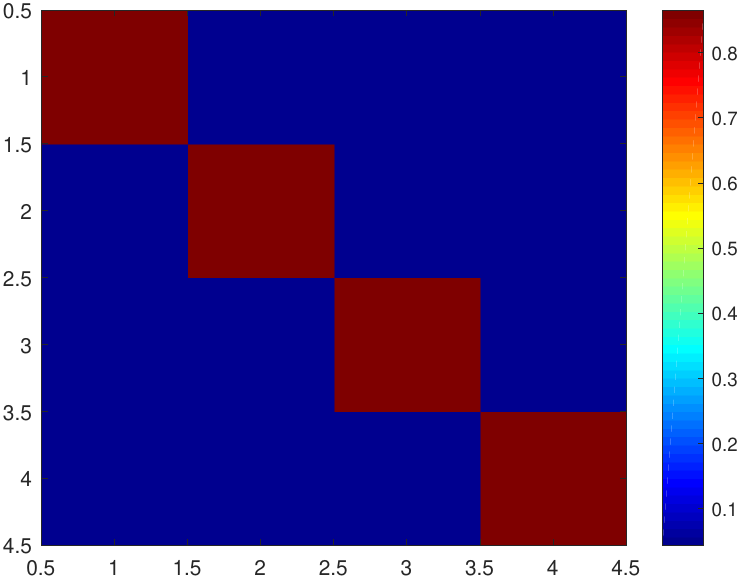}&
\includegraphics[width=1.65cm]{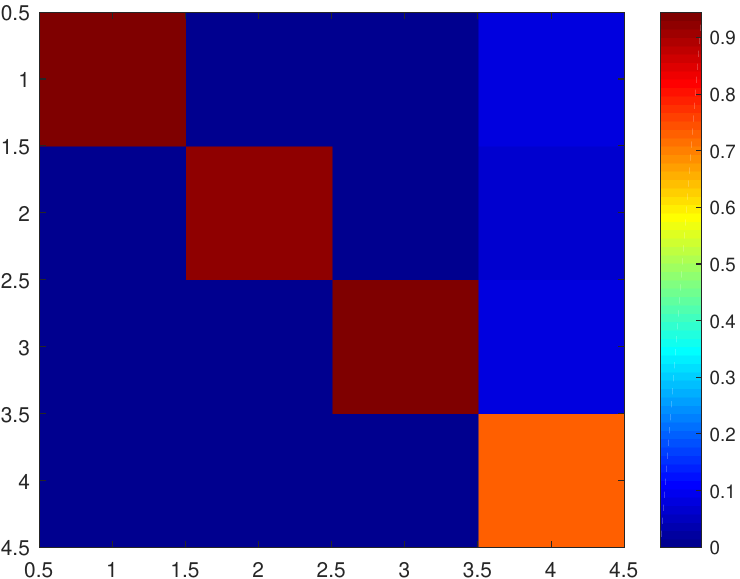}\\
\includegraphics[width=1.65cm]{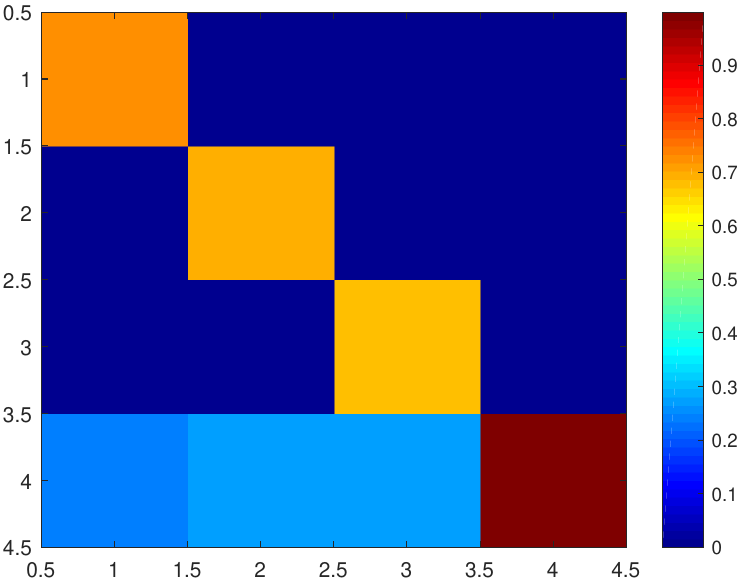}&
\includegraphics[width=1.65cm]{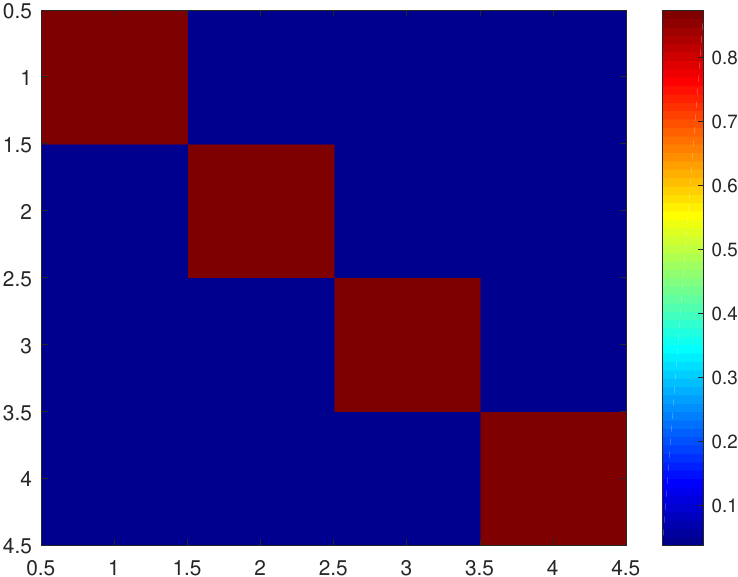}&
\includegraphics[width=1.65cm]{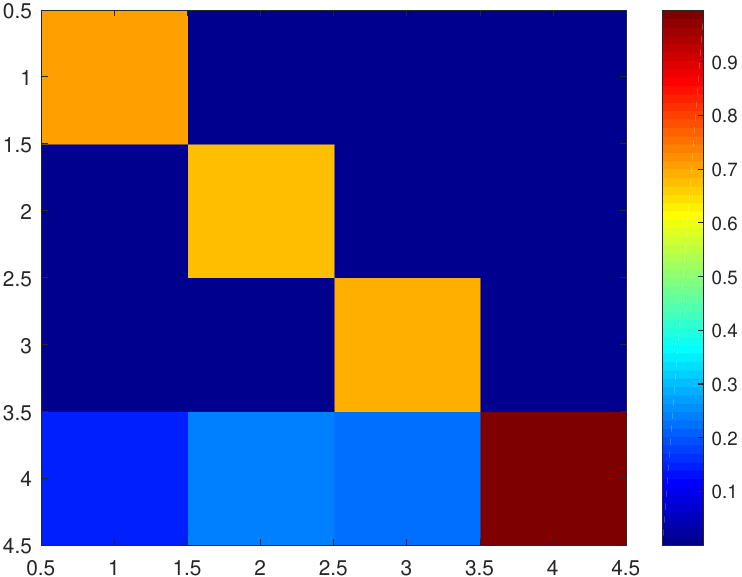}\\
\scriptsize{(a)} \hspace{0.0015in} &\scriptsize{(b)} \hspace{0.0015in} &\scriptsize{(c)} \hspace{0.0015in}
\end{tabular}
\vspace{0.2cm}
\caption{Comparison between the expected transition matrices (a), learned transition matrices by NTN (b), and learned average transition matrices by ANTN (c). The first and second rows represent the learned matrices for the erosion and dilation labels respectively.}
\end{wrapfigure}

\vspace{-0.1in}
\subsection{Experiments on histopathological images}

With the promising results with synthetic data, we further implement ANTN to DMD histo-images and compare segmentation results with both original K-Means and Otsu thresholding results and the results from previously evaluated deep-learning methods. 


It is difficult to obtain ground-truth pixel-by-pixel segmentation labels 
when studying histo-images in practice which essentially motivates the presented work as the existing deep-learning methods often rely on clean segmentation labels for model inference.
ANTN enables a new deep-learning model framework to incorporate noisy labels for training. For this set of experiments, we select 26 sub-images from one of 11 DMD histo-images with their corresponding K-Means and Otsu segmentation results as noisy segmentation labels. The example sub-images together with the corresponding segmentation results are shown in Figures 5(a), (b), and (c). As we observe empirically, K-Means often performs better than Otsu segmentation for our images. Model inference of u-net, NTN, and ANTN has been done similarly as for synthetic data. Note that u-net$^1$, u-net$^2$ and u-net$^3$ now represent the u-net trained with the corresponding K-Means, Otsu thresholding, and mixture of noisy segmentations. NTN$^1$ and NTN$^2$ represent the NTN trained with the corresponding K-Means and Otsu noisy segmentions. With the learning rate $10^{-4}$, training the u-net with the single type of noisy segmentations converges in 400 epochs and training with the mixture converges around 157 epochs. For NTN, we initialize the training with the corresponding u-net and then diffuse the noise transition layer by weight decay for 150 epochs. For ANTN, we initialize the clean label prediction network with the trained u-net$^3$ then further train two transition probability networks for 200 epochs. The consequent iterative adaptive training converges around 155 epochs with the same 10 epochs for the alternating interval as described earlier. 

We provide the corresponding segmentation results from u-net, NTN, and ANTN in Figures 5(d)-(i), which visually demonstrates that ANTN achieves the most homogeneous and coherent segmentations of fibrosis, muscle, and other tissue type regions based on the stains. Without ground-truth segmentation, we further evaluate the ratio of uniformity within segmented region to disparity aross segmented region of the original image intensities in $L*a*b*$ space as suggested in~\cite{chen2004use, zhang2008image}, and the smaller the ratio is, the better the segmentation is. The performance comparison results for all 11 original histo-images are given in Table 1 and
the entries within one standard deviation from the best segmentation results are highlighted in the table. Using the mixtures of K-Means and Otsu segmentation as training labels, u-net$^3$ and ANTN obtain better segmentation results compared to the other methods overall. This shows that integrating different types of noisy labels for deep network model training may help improve the performance. More importantly, ANTN achieves the best segmentation performance for 9 of 11 images.
\begin{figure}[h]
\centering
\begin{tabular}{@{}c@{}c@{}c@{}c@{}c@{}c@{}c@{}c@{}c@{}}
\includegraphics[width=1.35cm]{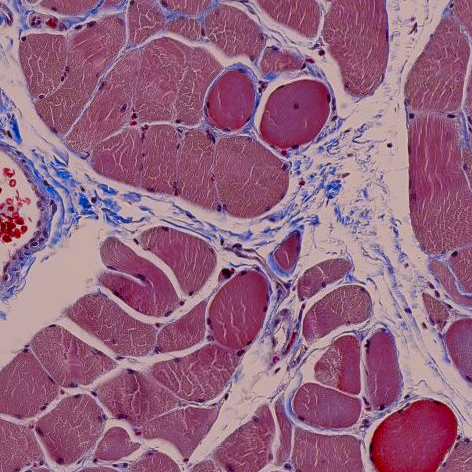}&\hspace{0.0015in}
\includegraphics[width=1.35cm]{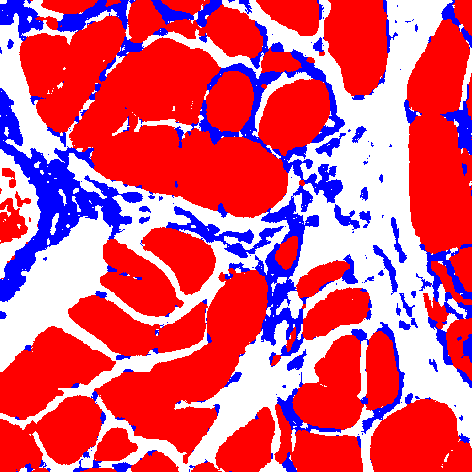}&\hspace{0.0015in}
\includegraphics[width=1.35cm]{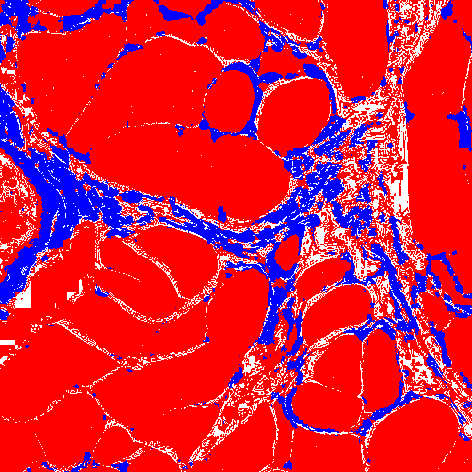}&\hspace{0.0015in}
\includegraphics[width=1.35cm]{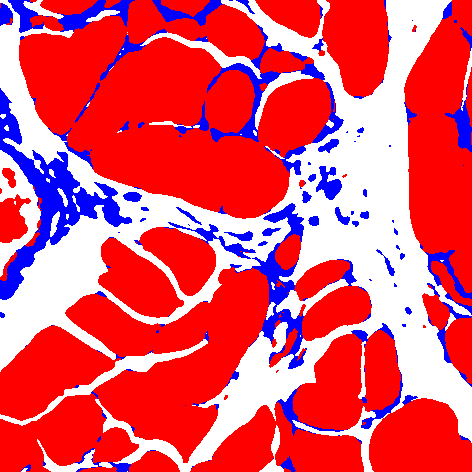}&\hspace{0.0015in}
\includegraphics[width=1.35cm]{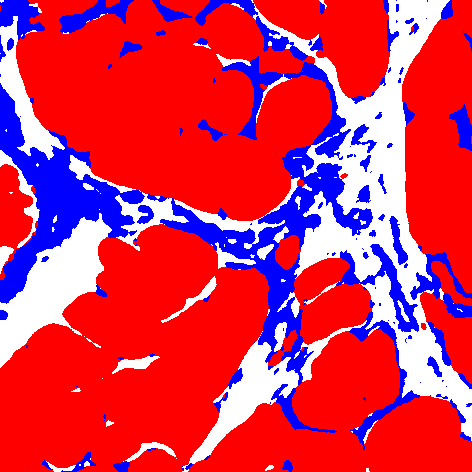}&\hspace{0.0015in}
\includegraphics[width=1.35cm]{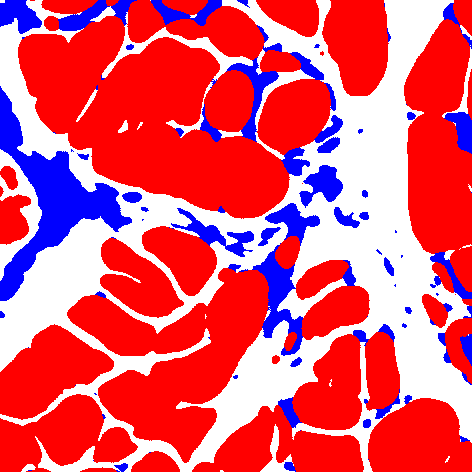}&\hspace{0.0015in}
\includegraphics[width=1.35cm]{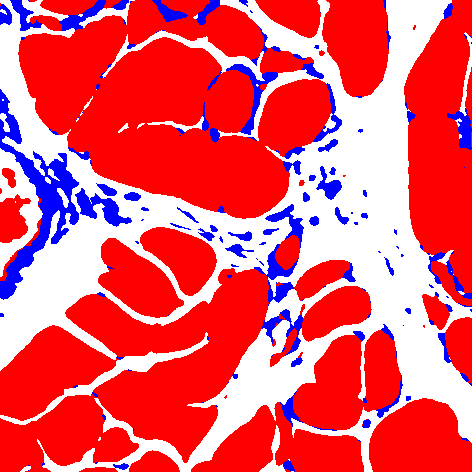}&\hspace{0.0015in}
\includegraphics[width=1.35cm]{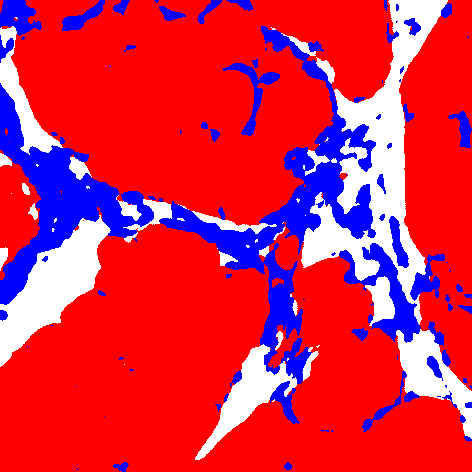}&\hspace{0.0015in}
\includegraphics[width=1.35cm]{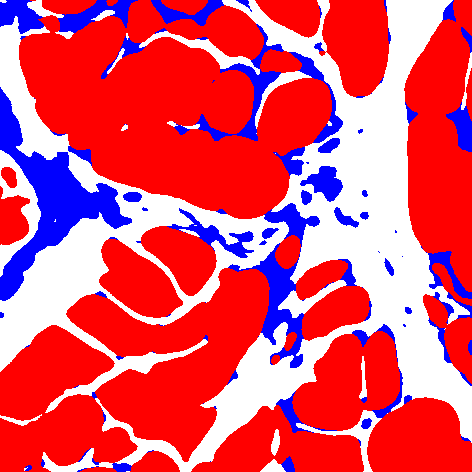}\\
\includegraphics[width=1.35cm]{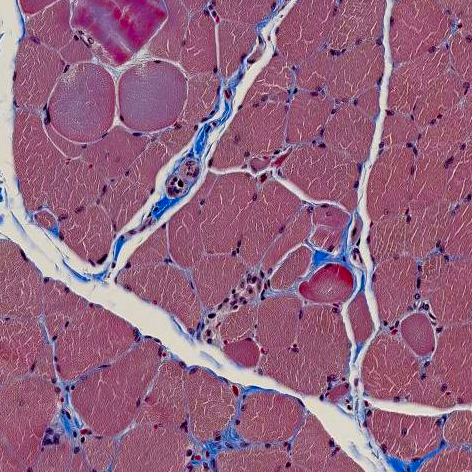}&\hspace{0.0015in}
\includegraphics[width=1.35cm]{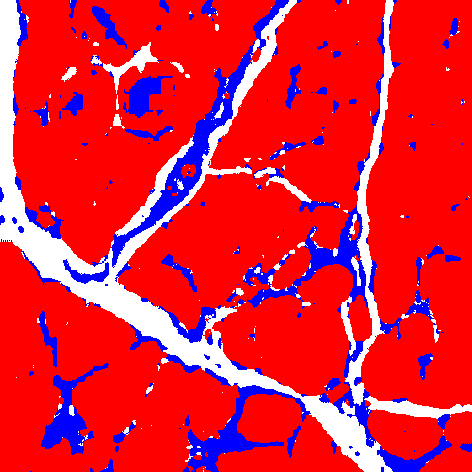}&\hspace{0.0015in}
\includegraphics[width=1.35cm]{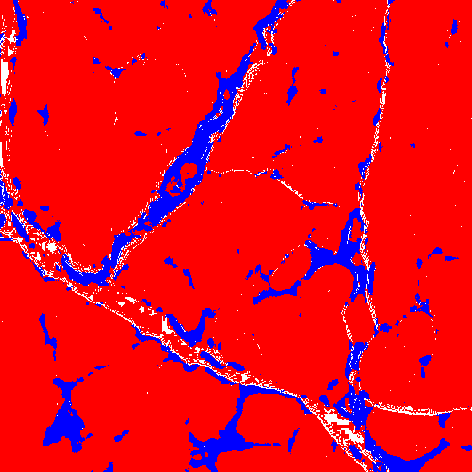}&\hspace{0.0015in}
\includegraphics[width=1.35cm]{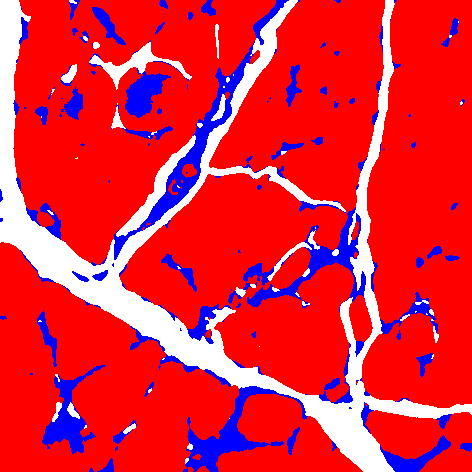}&\hspace{0.0015in}
\includegraphics[width=1.35cm]{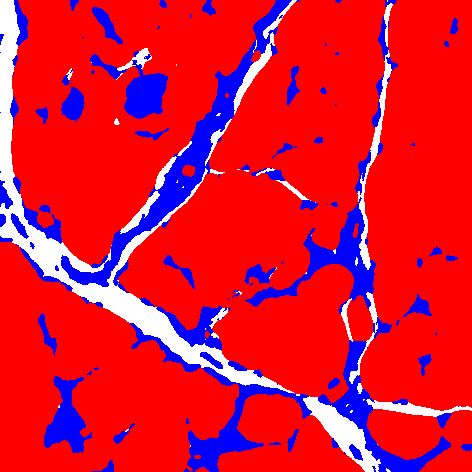}&\hspace{0.0015in}
\includegraphics[width=1.35cm]{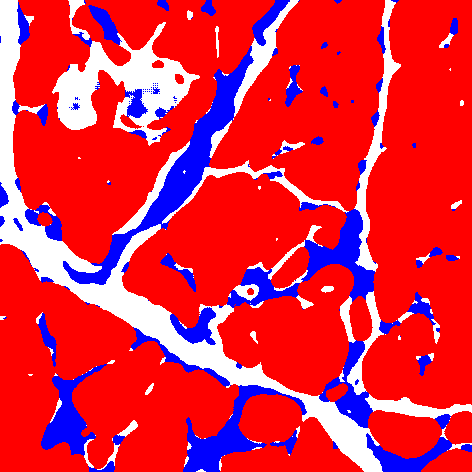}&\hspace{0.0015in}
\includegraphics[width=1.35cm]{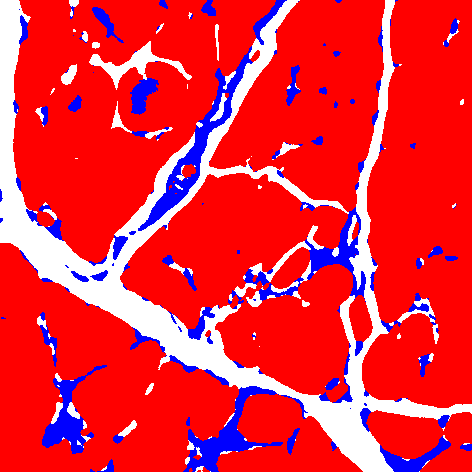}&\hspace{0.0015in}
\includegraphics[width=1.35cm]{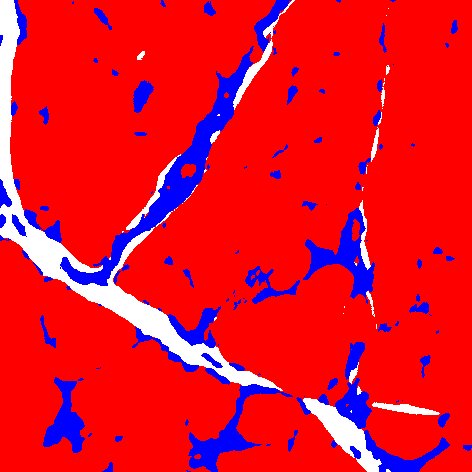}&\hspace{0.0015in}
\includegraphics[width=1.35cm]{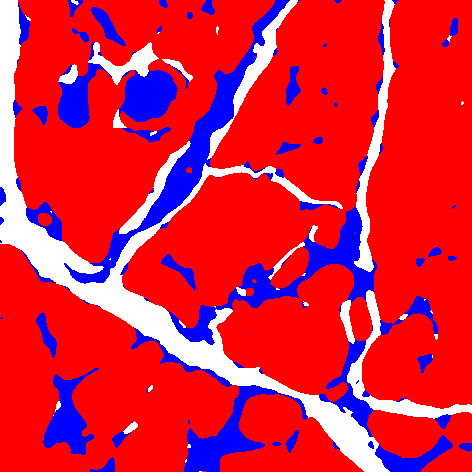}\\
\includegraphics[width=1.35cm]{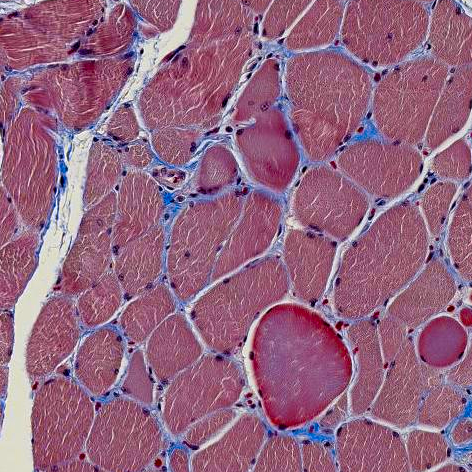}&\hspace{0.0015in}
\includegraphics[width=1.35cm]{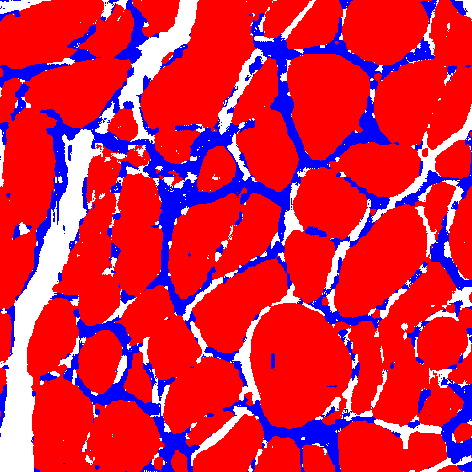}&\hspace{0.0015in}
\includegraphics[width=1.35cm]{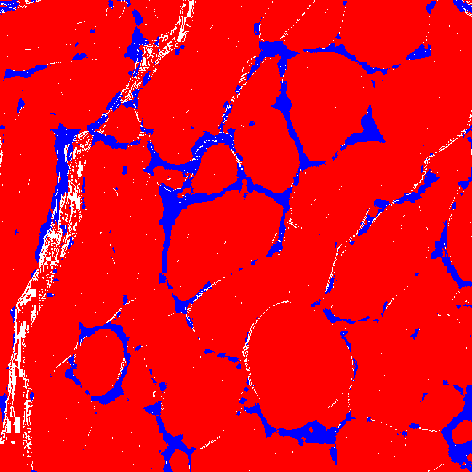}&\hspace{0.0015in}
\includegraphics[width=1.35cm]{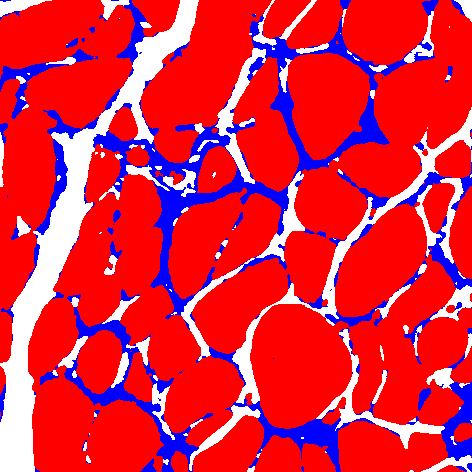}&\hspace{0.0015in}
\includegraphics[width=1.35cm]{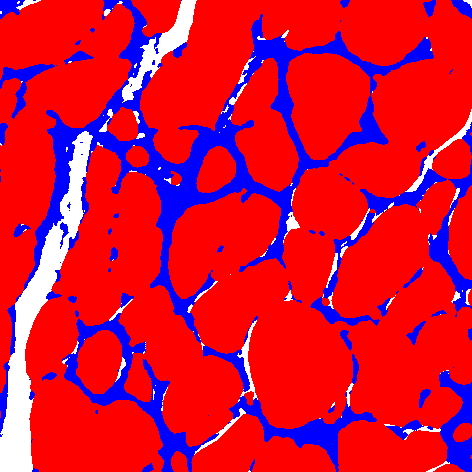}&\hspace{0.0015in}
\includegraphics[width=1.35cm]{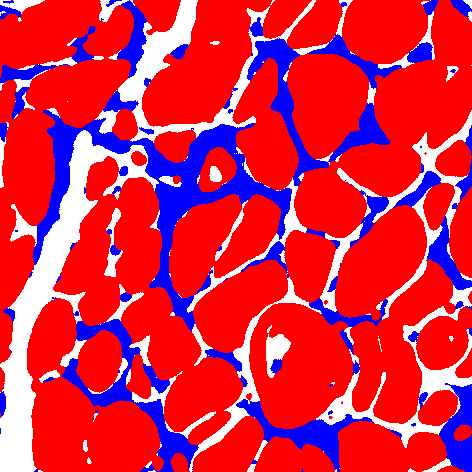}&\hspace{0.0015in}
\includegraphics[width=1.35cm]{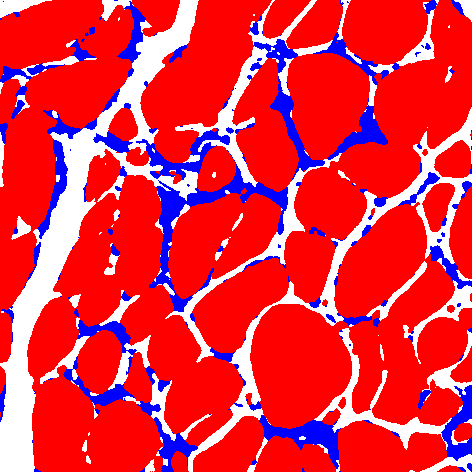}&\hspace{0.0015in}
\includegraphics[width=1.35cm]{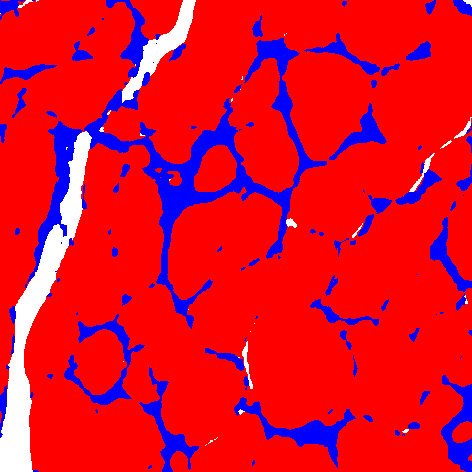}&\hspace{0.0015in}
\includegraphics[width=1.35cm]{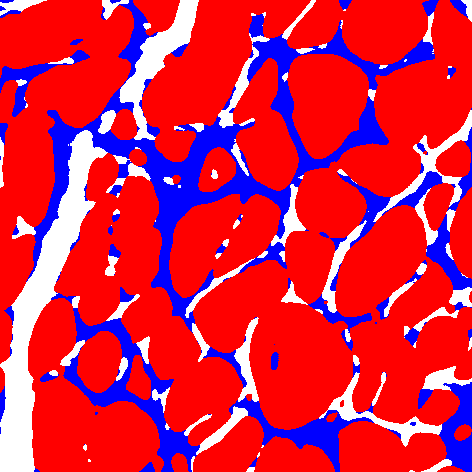}\\
\scriptsize{(a) Original} \hspace{0.0015in} & \scriptsize{(b) K-Means} \hspace{0.0015in} & \scriptsize{(c) Otsu} \hspace{0.0015in} & \scriptsize{(d) U-net$^1$} \hspace{0.0015in} & \scriptsize{(e) U-net$^2$} \hspace{0.0015in} & \scriptsize{(f) U-net$^3$} \hspace{0.0015in} & \scriptsize{(g) NTN$^1$} \hspace{0.0015in} & \scriptsize{(h) NTN$^2$} \hspace{0.0015in} & \scriptsize{(i) ANTN}
\end{tabular}
\vspace{0.2cm}
\caption{Original histo-images and corresponding segmentation results}
\end {figure} 
\begin{table}[h]
\vspace{0.3cm}
\centering
\label{my-label}
\scalebox{0.7}{
\begin{tabular}{|c|c|c|c|c|c|c|c|c|c|c|c|}
\hline
Method & 1      & 2      & 3      & 4      & 5      & 6      & 7      & 8      & 9      & 10     & 11     \\ \hline
K-Means                   & 0.3501 & 0.6183 & \textbf{0.2594} & 0.3432 & 0.2748 & \textbf{0.2177} & 0.6241 & 0.4196 & 0.4211 & 0.5025 & \textbf{0.2335} \\ \hline
Otsu                      & 0.3158 & 0.4928 & 0.3050  & 0.3529 & 0.3129 & 0.2558 & 0.5480  & 0.3653 & 0.4219 & 0.4995 & \textbf{0.2502} \\ \hline
U-net$^1$ & \textbf{0.2854} & 0.5123 & \textbf{0.2429} & 0.3254 & \textbf{0.2444} & \textbf{0.2271} & 0.4847 & 0.3506 & 0.3827 & 0.4742 & \textbf{0.1603} \\ \hline
U-net$^2$ & \textbf{0.2940} & 0.5058 & \textbf{0.2580} & 0.3233 & \textbf{0.2504} & 0.2308 & 0.5018 & 0.3505 & 0.3932 & 0.5152 & \textbf{0.1959} \\ \hline
U-net$^3$ & 0.3150  & \textbf{0.2917} & 0.2831 & \textbf{0.2810}  & \textbf{0.2578} & 0.2467 & \textbf{0.2898} & 0.3424 & \textbf{0.2709} & \textbf{0.3036} & 0.7437 \\ \hline
NTN$^1$  & \textbf{0.2848} & 0.4978 & \textbf{0.2484} & 0.3239 & \textbf{0.2470}  & \textbf{0.2280}  & 0.4955 & 0.3520  & 0.3857 & 0.4823 & \textbf{0.1594} \\ \hline
NTN$^2$  & 0.3128 & 0.5066 & 0.2861 & 0.3330  & 0.2737 & 0.2473 & 0.5225 & 0.3584 & 0.4213 & 0.5500   & \textbf{0.2281} \\ \hline
ANTN  & \textbf{0.2751} & \textbf{0.2790} & 0.2676 & \textbf{0.2663} & \textbf{0.2472} & 0.2332 & \textbf{0.2788} & \textbf{0.3113} & \textbf{0.2670}  & \textbf{0.2966} & \textbf{0.2311} \\ \hline
\end{tabular}}
\vspace{0.2cm}
\caption{Performance comparison of different methods on 11 original histo-images.}
\end{table}

We also investigate the estimated ratio similarly as for synthetic data based on the intermediate outputs during ANTN training by noisy segmentations from either K-Means or Otsu algorithm. It is observed that the ratio with respect to K-Means is much larger than that with Otsu (Figure 6(a)). Besides, the corresponding average transition matrices after convergence are shown in Figure 6(b) and (c). Clearly, the average label-flip noise transition matrix trained for K-Means segmentation has diagonal entry values closer to 1 compared to that for Otsu segementation. This tells that K-Means segmentation results match better with the segmentation results derived by ANTN, indicating that K-Means achieves better segmentation results compared to Otsu thresholding. This again agrees with our empirical observation from the beginning.

\begin{figure}[h]
\centering
\vspace{-0.1in}
\begin{minipage}[\vtop]{3.9cm}
\includegraphics[width=3.9cm]{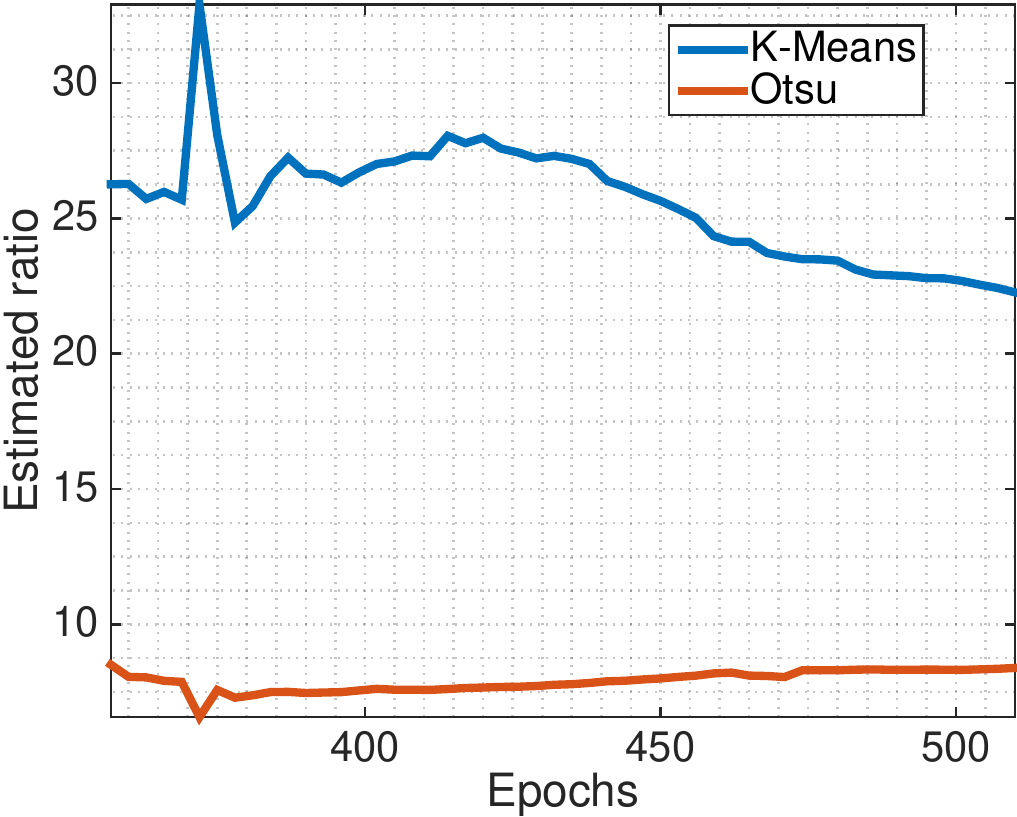}\\\centering\scriptsize{(a)}
\end{minipage}
\begin{minipage}[\vtop]{1.65cm}
\includegraphics[width=1.65cm]{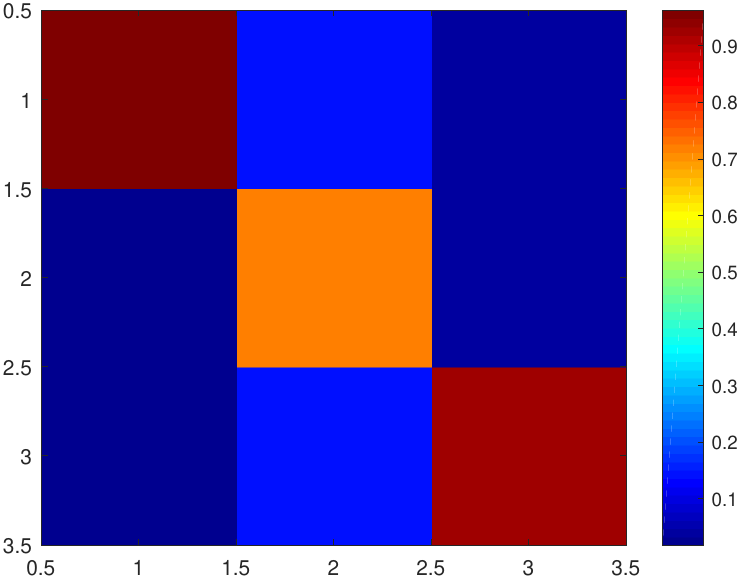}\\\centering\scriptsize{(b)}\vspace{0.1cm}
\includegraphics[width=1.65cm]{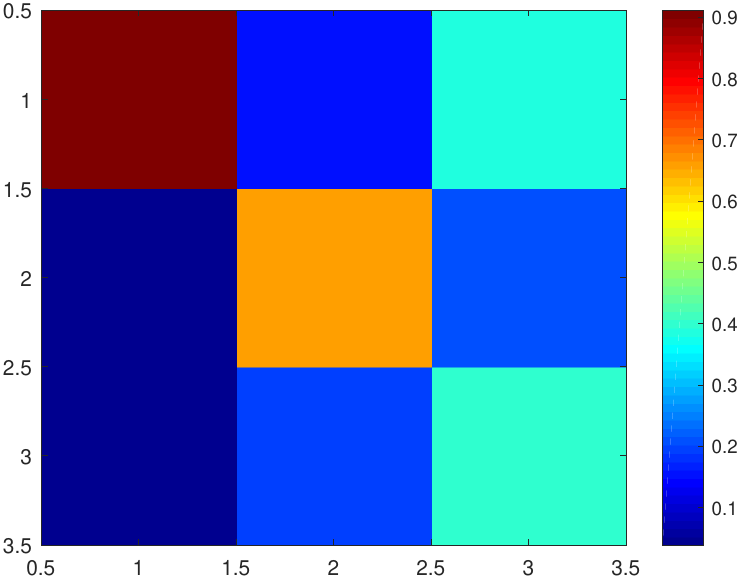}\\\centering\scriptsize{(c)}
\end{minipage}
\vspace{0.2cm}
\caption{(a) Estimated ratio of intermediate output labels to noisy labels. (b) Estimated transition matrix for K-Means noisy dataset. (c) Estimated transition matrix for Otsu noisy dataset.}
\end{figure}

\vspace{-0.1in}
\section{Conclusions}

We have proposed a novel adaptive noise-tolerant network (ANTN) to integrate multiple noisy datasets for image segmentation. ANTN models the feature-dependent transition probabilities adaptively from multiple off-the-shelf  segmentation algorithms that help generate noisy labels for training. Based on the extensive performance evaluation on both synthetic data and real-world histo-images, it is clear that ANTN is a promising automated deep-learning image segmentation method that can take noisy or ``weak'' segmentation results and further improve segmentation performance by borrowing signal strengths from multiple weak labels. 

\bibliographystyle{IEEEtran}
\bibliography{egbib}
\end{document}